\documentclass[conference]{IEEEtran}
\IEEEoverridecommandlockouts
\usepackage{cite}
\usepackage{amsmath,amssymb,amsfonts}
\usepackage{algorithmic}
\usepackage{graphicx}
\usepackage{textcomp}
\usepackage{xcolor}
\usepackage{booktabs} 
\usepackage[colorlinks=false,pdfborder={0 0 0}]{hyperref}
\usepackage[utf8]{inputenc}

\def\BibTeX{{\rm B\kern-.05em{\sc i\kern-.025em b}\kern-.08em
    T\kern-.1667em\lower.7ex\hbox{E}\kern-.125emX}}

\begin{document}

\title{Cross-Granularity Representations for Biological Sequences: Insights from ESM and BiGCARP
\\
\thanks{This work was supported by the Engineering and Physical Sciences Research Council [EP/Y030826/1].}
\thanks{\textcopyright{} 2025 IEEE. Personal use of this material is permitted. Permission from IEEE must be obtained for all other uses, in any current or future media, including reprinting/republishing this material for advertising or promotional purposes, creating new collective works, for resale or redistribution to servers or lists, or reuse of any copyrighted component of this work in other works. Published in: 2025 IEEE International Conference on Bioinformatics and Biomedicine (BIBM), pp. 6936--6943, DOI: 10.1109/BIBM66473.2025.11356265}
}

\author{\IEEEauthorblockN{
Hanlin Xiao\(^{1,2}\),
Rainer Breitling\(^{1,*}\),
Eriko Takano\(^{1}\),
Mauricio A. Alvarez\(^{2}\)}
\IEEEauthorblockA{
\(^{1}\) Manchester Institute of Biotechnology, The University of Manchester, Manchester, United Kingdom}

\IEEEauthorblockA{
\(^{2}\) Department of Computer Science, The University of Manchester, Manchester, United Kingdom}

\IEEEauthorblockA{
* Corresponding author: Rainer.Breitling@manchester.ac.uk}}

\maketitle

\begin{abstract}
Recent advances in general-purpose foundation models have stimulated the development of large biological sequence models. While natural language shows symbolic granularity (characters, words, sentences), biological sequences exhibit hierarchical granularity whose levels (nucleotides, amino acids, protein domains, genes) further encode biologically functional information.
In this paper, we investigate the integration of cross-granularity knowledge from models through a case study of BiGCARP, a Pfam domain–level model for biosynthetic gene clusters, and ESM, an amino acid–level protein language model. Using representation analysis tools and a set of probe tasks, we first explain why a straightforward cross-model embedding initialization fails to improve downstream performance in BiGCARP, and show that deeper-layer embeddings capture a more contextual and faithful representation of the model's learned knowledge. Furthermore, we demonstrate that representations at different granularities encode complementary biological knowledge, and that combining them yields measurable performance gains in intermediate-level prediction tasks. Our findings highlight cross-granularity integration as a promising strategy for improving both the performance and interpretability of biological foundation models. Our code is available at \url{https://github.com/Nugkta/cgrep}.
\end{abstract}

\section{Introduction}

\begin{figure}[!h]
    \centering    
    \includegraphics[width=1\linewidth]{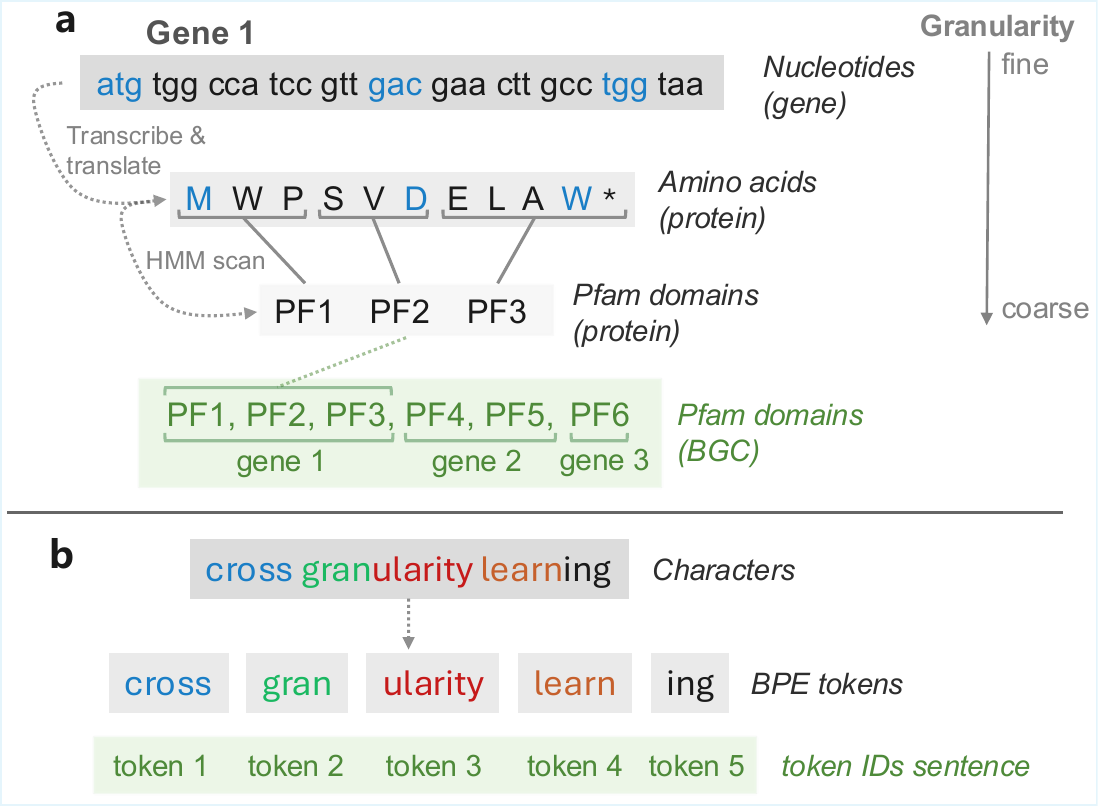}
    \caption{\textbf{Multi-granularity nature of biological sequences and its analogy to tokenization in NLP.} a) Gene sequences at the nucleotide level can be transcribed and translated into amino acids, which can be further abstracted into Pfam domains by HMM scans. A biosynthetic gene cluster (BGC) can thus be represented as a sequence of Pfam domains, the granularity used in the BiGCARP model. b) Similarly, NLP sequences can be represented at different granularities, from characters through subwords (the ``token'' here) to words and sentences. Unlike statistical NLP tokenizers (e.g., byte-pair encoding), which are reversible, biological “tokenizers” are lossy but inject functional priors by providing higher-level abstractions.}
    \label{fig:CrossGran}
\end{figure}

Recent breakthroughs in deep learning~\cite{devlin2018bert, lecun2015deep} have spurred a paradigm shift towards the training of foundation models \cite{bommasani2021opportunities}. These models are characterized by their general-purpose architectures, trained on vast, multi-domain datasets using self-supervised learning frameworks. Given the complexity and data-rich nature of biology, the field has witnessed a surge of new biological foundation models spanning multiple modalities and domains, such as genomics, medical imaging, and transcriptomics \cite{guo2025foundation}. Similar to text in natural language, biological sequences, such as the nucleotide sequences of DNA and the amino acid sequences of proteins, are central to this revolution. Inspired by the remarkable success of Large Language Models (LLMs) \cite{zhao2023survey}, biological sequence models aim to learn the underlying ``grammar'' of these sequences. The recent proliferation of such models \cite{brixi2025genome,zhou2023dnabert,hayes2025simulating} and their applications has demonstrated their significant effectiveness.

A defining aspect of biological sequences is their intrinsic hierarchical and \textit{multi-granular} nature, which offers an analogy to tokenization in natural language processing, albeit with vital distinctions (Fig.~\ref{fig:CrossGran}). Whereas NLP tokenizers (e.g., Byte Pair Encoding) are typically lossless, data-driven, and optimized from a corpus to form an efficient vocabulary \cite{devlin2018bert}, transitions between biological scales are governed by biophysical principles (Fig.~\ref{fig:CrossGran}a) and inherently lossy. For instance, the translation from nucleotides to amino acids discards information of non-coding regions and the use of synonymous codons. On the other hand, this abstraction exchanges genomic complexity for a representation where functional and structural information, such as conserved protein motifs, becomes more explicit. This principle of abstraction extends to higher levels. A full protein sequence can be represented as an ordered sequence of its constituent functional domains, often defined by models from databases like Pfam \cite{mistry2021pfam}. This transformation is again lossy, but it distills a complex polypeptide into its core functional and evolutionary units, which immediately provides functional hypotheses and simplifies comparisons of protein architecture across evolutionary distances. At the highest level relevant here, this domain-centric view can be applied to an entire biosynthetic gene cluster (BGC), which is a cluster of co-located genes that cooperate to produce a secondary metabolite. As shown in Fig. \ref{fig:CrossGran}, a BGC can be represented as a sequence of domains, ignoring the boundaries between individual proteins. This provides a simplified representation of the enzymatic ``assembly line,'' which could act as a powerful heuristic for predicting the chemical class of the synthesized metabolites.

This multi-granular nature is mirrored by the distinct types of information prioritized by databases at each level. Genomic databases capture the whole genomic context, annotating features including gene locations, regulatory elements, and non-coding regions. In contrast, protein-centric databases, even when computationally annotated, are structured to store protein-specific functional and structural information, such as predicted three-dimensional structures and Gene Ontology (GO) term annotations. Therefore, foundation models have typically been developed to specialize at a single granularity---from the nucleotide \cite{zhou2023dnabert, brixi2025genome} and amino acid levels \cite{brandes2022proteinbert, brandes2023genome}, to the gene or domain level \cite{hwang2024genomic}. Each model leverages the specific information available at its chosen scale, allowing coarser models to capture long-range dependencies while finer models resolve local details, ultimately leading to highly specialized and distinct representation spaces.

\begin{figure}[htbp]
    \centering
    \includegraphics[width=0.8\linewidth]{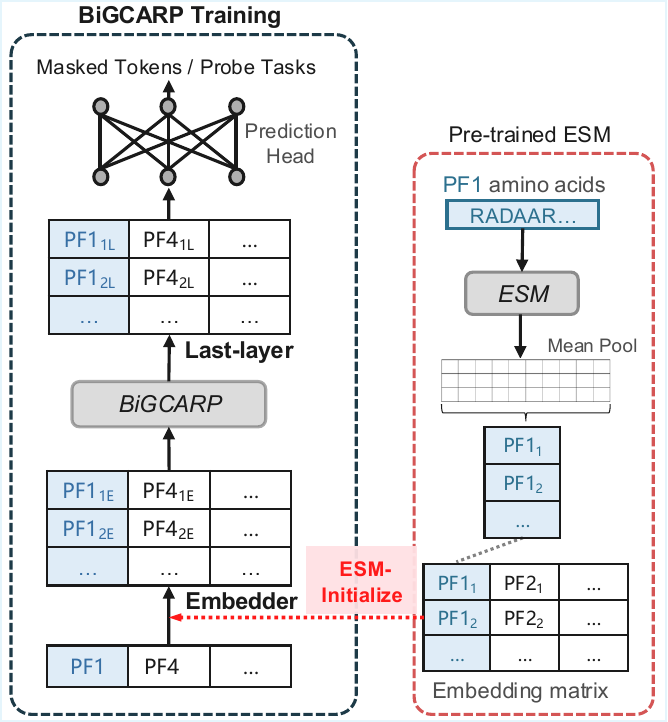}
    \caption{
    \textbf{Overview of the BiGCARP model and ESM-based initialization.} 
    \textbf{Left:} A BGC is represented as a sequence of Pfam domains, 
    which are embedded through an embedding matrix initialized either randomly or with ESM. 
    BiGCARP processes these embeddings into last-layer representations, which are passed to a prediction head 
    for masked language modeling and downstream probe tasks. 
    \textbf{Right:} In ESM-based initialization, each Pfam domain is associated with a representative amino acid sequence. 
    The sequence is encoded by the pre-trained ESM model, mean-pooled, and used to construct the embedding vector 
    for that domain in BiGCARP.
    \textbf{Blue blocks} illustrate an example of domain PF1 receiving its embedding initialization 
    $\{\text{PF}_{i}\}_{i=1}^n$ from its representative amino acid sequence (right), 
    and progressing from its embedder-layer embeddings 
    $\{\text{PF}_{iE}\}_{i=1}^n$ to its last-layer embeddings $\{\text{PF}_{iL}\}_{i=1}^n$ (left). 
}
    \label{fig:EsmInit}
\end{figure}

This paper investigates the potential of integrating information across these distinct granularities through an empirical case study of two representative models: ESM
\cite{hayes2025simulating}, and BiGCARP \cite{rios2023deep}. ESM is a series of massive amino acid-level models; the most recent version is pre-trained on over 3.15 billion protein sequences, capturing broad biological domains. It has demonstrated state-of-the-art performance on diverse downstream tasks, including protein structure prediction and de novo sequence generation. In contrast, BiGCARP operates at a much coarser granularity of Pfam domains \cite{mistry2021pfam} to manage the longer context length of BGCs. It is trained on the more specialized antiSMASH database \cite{blin2021antismash}, which contains around 127,000 BGCs. Furthermore, the original BiGCARP implementation has already attempted to leverage ESM's representations by initializing BiGCARP's token embedding matrix using domain-level representations derived from ESM (Fig. \ref{fig:EsmInit}). However, despite the seemingly informative initialization, the strategy did not translate to a substantial improvement in downstream task performance. This discrepancy between the potential of the cross-granular information and the outcome presents another question that our research addresses.

The main contributions of this paper include:
\begin{itemize}
    \item We analyze the training dynamics of the BiGCARP model to explain why its ESM-based initialization does not transfer effectively. We demonstrate that the model's initial architectural layers largely transform the representation space, preventing the initialized information from being retained in higher layers.
    \item We investigate the choice of embedding extraction strategy, showing that representations from the last layer are more faithful to the model's learned knowledge and result in better downstream performance in this cross-granularity context.
    \item By curating three evaluation tasks, we demonstrate that fine-grained (ESM) and coarse-grained (BiGCARP) models learn complementary local and global information. We further show that combining these representations improves performance for an intermediate-level task, underscoring cross-granularity integration as a promising approach for developing more comprehensive biological foundation models.
\end{itemize}

\section{Related Work}
Multimodal foundation models in biology have recently gained traction, with many approaches integrating heterogeneous data types such as structural information, textual descriptions of function, images, or omics profiles \cite{guo2025foundation, lin2023evolutionary}. A range of biological sequence models have been developed across different granularities, such as nucleotide-based models \cite{zhou2023dnabert, brixi2025genome}, amino acid models \cite{brandes2022proteinbert, brandes2023genome}, gene-level models  \cite{hwang2024genomic} and Pfam domain-level models like BiGCARP \cite{rios2023deep}. However, these models are usually trained and evaluated independently, with little effort focused on fusing cross-granular representations to obtain complementary insights.
Early efforts in cross-granularity integration have been explored in natural language processing (NLP) \cite{kim2024initializing}, where BERT was initialized with pre-trained embeddings from other models. Also, in gLM \cite{hwang2024genomic}, the authors adopted similar initialization strategies at the gene level. Finally, several task-specific applications relate to our experiments: for example, \cite{liu2025unified} uses BiGCARP’s embedder representations to match BGCs and their products, and existing A-domain prediction methods \cite{adduri2025interpretable} often adopt ESM-based embeddings---a strategy we adapt and extend in this study.

\section{Methods and Experimental Design}
    \subsection{Relevant Biological Sequence Models}

        \subsubsection{ESM}
        The Evolutionary Scale Modeling (ESM) family consists of powerful protein language models based on encoder-only Transformer architectures. The most recent model, ESM-3, is trained on over 3.15 billion protein sequences, including synthetic data, and incorporates multimodal inputs such as structural and functional annotations \cite{hayes2025simulating}. Since this project focuses on prokaryotic BGCs and sequence-level modeling, we primarily employ ESM-2 and ESM-1b to maintain consistency with the original BiGCARP pipeline. Given an amino acid sequence, these models output contextual embeddings for each residue; in our case, the embeddings are 1,280-dimensional. For full architectural and pre-training details, we refer readers to the original ESM publications \cite{lin2023evolutionary}.

        \subsubsection{BiGCARP}

        BiGCARP also adopts an encoder-based architecture, but is built upon the CARP framework \cite{yang2024convolutions}, which leverages dilated convolutional layers for efficient sequence modeling. Unlike ESM, BiGCARP is specifically designed for BGC analysis. It is pre-trained on data from the antiSMASH database \cite{blin2021antismash}, which contains approximately 127,000 annotated BGCs, including nucleotide sequences, gene annotations, and translated amino acid sequences.

        During pre-training, each BGC is scanned using Hidden Markov Models (HMMs) from the Pfam database to generate a sequence of protein domains. These domain sequences typically range from a few dozen to several hundred Pfam entries per BGC. BiGCARP takes these domain sequences as input and produces embeddings for each Pfam domain. 
        \subsection{The Cross-Granularity Initialization}
        The pipeline for ESM-based embedding initialization is illustrated in Fig. \ref{fig:EsmInit}. In the standard BiGCARP model, each Pfam domain in the input sequence is first mapped to a representative vector using an embedding matrix. Usually, this matrix is randomly initialized from a Gaussian distribution, following standard practice in language modeling (e.g., BERT), and jointly optimized with the model during training.

        However, to incorporate information from the finer-grained, pre-trained ESM model, BiGCARP also initializes the Pfam embedding matrix using ESM-derived representations. For each Pfam domain, a representative amino acid sequence is selected (the first entry associated with that domain in the Pfam database) and passed through ESM. The last-layer residue embeddings are then mean-pooled to obtain a fixed-size vector, which is used to initialize the corresponding entry in the embedding matrix, as shown on the right side of Fig. \ref{fig:EsmInit}.
        
    \subsection{Embedding Extraction}

    \textit{Embeddings} are vector representations of tokens in the vocabulary, designed to encode information learned during pre-training. These embeddings serve as input to downstream modules, as illustrated in the top-left corner of Fig. \ref{fig:EsmInit}.
    In the original BiGCARP implementation, embeddings are extracted directly from the embedding matrix layer, resulting in static representations---each Pfam domain is always associated with the same vector. These embeddings are thus strongly influenced by the initialization method, such as the ESM-derived scheme described earlier.
    
    However, in representation learning, it is more common to extract embeddings from higher layers of the model. These layers incorporate contextual information through mechanisms like self-attention or dilated convolutions, enabling the model to produce contextualized representations, where the same token may have different embeddings depending on its surrounding domains. For example, in BERT, representation extraction is often performed from higher or final layers \cite{devlin2018bert}. Moreover, different layers have been shown to capture distinct types of linguistic and semantic information \cite{rogers2021primer}. Motivated by this, we compare the downstream task performance using embeddings extracted from the initial embedding matrix (\textit{embedder layer}) versus the \textit{last-layer embeddings} of BiGCARP (Fig. \ref{fig:EsmInit}).
    
    Additionally, for our UMAP-based representation analysis, we require a single, static embedding per Pfam domain. To construct this, we aggregate all occurrences of each Pfam domain across the dataset, extract their contextualized embeddings, and compute their average. These averaged vectors are then used for UMAP dimensionality reduction and visualization.

    \subsection{Representation Analysis Tools}

        \subsubsection{Uniform Manifold Approximation and Projection (UMAP)}
        To analyze the representations learned by each model, we apply UMAP, a dimensionality reduction method that preserves both local and global structure \cite{mcinnes2018umap}. High-dimensional embeddings, 1280-dimensional for ESM and 256-dimensional for BiGCARP, are projected into two dimensions for visualization. Each Pfam domain is associated with a representative embedding, visualized as a point in 2D. We select UMAP over other dimension reduction methods, such as PCA, due to its stronger preservation of the representation space structure. A meaningful embedding space should result in reduced representations that exhibit clustering of similar Pfam domains.

        To interpret the resulting visualizations, we apply several labeling strategies. Following the original BiGCARP paper, each domain is first labeled by its Pfam clan, which captures sequence similarity and evolutionary origin \cite{finn2006pfam}. We further annotate each domain with one of four functional categories---core, tailoring, regulation, or transport---based on its role within a BGC. Finally, we manually examine representative clusters to assess whether they correspond to biologically coherent groupings, offering insight into the functional information encoded by the embeddings.

        \subsubsection{Centered Kernel Alignment (CKA)}
        CKA is a technique for quantifying the similarity between two sets of representations. It computes a normalized similarity score by comparing how patterns of variation in one representation space correspond to those in another, effectively measuring how similarly the two spaces organize the same set of inputs \cite{kornblith2019similarity}.

        In this study, we use CKA to analyze representational dynamics in the BiGCARP model under different initialization schemes. Specifically, we compare representations across layers to assess how information evolves as it propagates through the network. We compute CKA scores either between layers of the same model or across models with different initialization (e.g., ESM-based vs. random). Higher CKA scores indicate greater representational similarity, which we interpret as a proxy for how much information is preserved during encoding.

    \subsection{Probe Tasks}
    To compare the effectiveness of different embeddings, we select three probe tasks adapted from prior studies. These tasks serve as targeted evaluations of how well various representations capture features of different scales. This section introduces the biological motivation behind each task and describes the corresponding datasets. For the performance metrics, we have mainly focused on the comparison of the Macro AUROC score, as it provides a robust, threshold-independent measure of class separability and is particularly suitable for imbalanced classification tasks.

        \subsubsection{BGC Product Class Prediction}
        We evaluate embedding quality through a BGC product class prediction task, following prior work \cite{hannigan2019deep, rios2023deep}. Each BGC is linked to a characterized natural product, categorized into one or more of seven major classes: polyketides, nonribosomal peptides (NRPs), RiPPs, terpenes, saccharides, alkaloids, and others. This task is framed as a multi-label classification problem.

        These product classes differ in their biosynthetic logic. For instance, polyketides are synthesized by polyketide synthases (PKSs) through iterative or modular extension using acyl-CoA precursors, while NRPs are assembled by nonribosomal peptide synthetases (NRPSs) via domain-specific modules. Therefore, this task allows us to evaluate whether the embeddings capture more global biosynthetic patterns across the entire BGC.
        
        We focus on three key comparisons: (1) ESM-initialized vs. randomly initialized BiGCARP models, (2) last-layer (contextual) vs. embedder layer (static) representations, and (3) single-source vs. concatenated embeddings from ESM and BiGCARP. For consistency, we employ a simple two-layer MLP classifier (hidden dimensions: 512 and 256) across all evaluations. This architecture is sufficiently expressive to process high-dimensional embeddings while remaining deliberately limited in capacity, ensuring that performance differences reflect the quality of the embeddings rather than the classifier.
        
        Given the limited size (about 3,000 entries) and class imbalance of the training dataset, we adopt stratified 5-fold cross-validation to ensure balanced splits. To improve statistical robustness, each evaluation is repeated across 10 random seeds, and we report the average performance and uncertainties across runs.

        \subsubsection{A-Domain Substrate Prediction}

        Nonribosomal peptides (NRPs) are a class of natural products encoded by BGCs, with important applications as antibiotics, toxins, and immunosuppressants \cite{sussmuth2017nonribosomal}. They are assembled by nonribosomal peptide synthetases (NRPSs), which use amino acids as monomers to construct peptide backbones. NRPSs are organized into repeating modules, each typically containing three core domains: the \textit{adenylation (A) domain}, which selects and activates the substrate; the peptidyl carrier protein (PCP) domain, which tethers the activated amino acid; and the condensation (C) domain, which catalyzes peptide bond formation between adjacent residues.
        
        In this experiment, we focus on the A-domain, due to its central role in determining substrate specificity. Accurate prediction of A-domain substrates is essential for inferring NRP structure and function, but remains a challenging task. Several recent tools have employed deep learning and protein language models to tackle this problem~\cite{adduri2025interpretable, mongia2023adenpredictor}. Substrate specificity is largely governed by local sequence and structural features, particularly a well-characterized motif known as the \textit{Stachelhaus code}---a set of ten residues near the substrate-binding pocket~\cite{stachelhaus1999specificity}. In addition, the broader BGC context might influence substrate selection, for example, through epistatic effects or coordinated domain organization across modules, motivating the attempt at representation integration.
        
        To investigate this, we compare two embedding strategies:  
        (1) \textit{ESM-based embeddings}, derived from the A-domain amino acid sequence; and  
        (2) \textit{BiGCARP-based embeddings}, derived from the full Pfam-annotated BGC context.
        
        For the ESM variant, we input the full A-domain amino acid sequence into the ESM model and extract residue-level embeddings. We then isolate the embeddings corresponding to the Stachelhaus code positions and apply mean pooling to obtain the final representation. This strategy outperformed full-sequence pooling and is therefore the only variant reported here.
        
        For the BiGCARP variant, we input the full BGC Pfam domain sequences that contain the A-domain (PF00501) into the model and extract the last-layer embeddings. We then apply mean pooling over the entire BGC sequence to obtain a global representation. We also experimented with using only the contextualized A-domain embedding from BiGCARP, but found it performed worse than full-sequence pooling and thus omitted it from further analysis.
        
        The dataset used in this experiment is based on annotations introduced in~\cite{adduri2025interpretable}, which extend MIBiG entries with known A-domain substrates and Stachelhaus code residues. For each annotated A-domain, we retrieve its originating BGC, reconstruct the Pfam domain sequence, and align the A-domain position for embedding extraction. Out of approximately 2,200 annotated A-domains, around 1,600 could be successfully matched with corresponding BGC sequences.
        
        The substrate label space consists of roughly 50 distinct classes, many of which are sparsely populated. This long-tail distribution makes prediction particularly difficult and often necessitates specialized approaches such as contrastive learning using molecular structures annotations~\cite{adduri2025interpretable}. In this study, we adopt a simplified setup, predicting a set of key substrate-related properties as summarized in Table~\ref{tab:adomain}.

        \subsubsection{Halogen Product Prediction}
        This task targets the prediction of whether a BGC produces a halogenated natural product. We focus on Pfam domain PF04820, which corresponds to flavin-dependent halogenases that catalyze the incorporation of halogen atoms (e.g., Cl, Br) into organic compounds. While the presence of PF04820 suggests halogenation, this association is not deterministic: BGCs containing this domain do not always yield halogenated products due to factors such as gene silencing, incomplete biosynthetic context, or inactive domain variants.

        To investigate this, we curated a dataset from the MIBiG database, which includes manually annotated biosynthetic gene clusters along with the chemical structures of their known products. We filtered all BGCs containing PF04820 and labeled each as halogenated or non-halogenated based on whether any of its associated product SMILES strings contained halogen atoms (F, Cl, Br, or I).

        Due to the rarity of experimentally characterized halogenated BGCs, the resulting dataset comprises only 46 entries containing the PF04820 domain. To ensure statistical robustness despite the limited sample size, we adopt a leave-one-out cross-validation (LOOCV) scheme. Furthermore, we apply bootstrap resampling with 10,000 iterations to estimate confidence intervals for each embedding configuration, allowing for more reliable performance comparisons across models.

    \subsection{Datasets}
    
    \textbf{antiSMASH-DB}---This is the primary corpus used to train the BiGCARP model. It contains around 147,000 sequences, each processed using the antiSMASH tool to identify gene sequences and translate them into amino acid sequences. To reduce redundancy and prevent data leakage, preprocessing follows the original BiGCARP protocol, including BiG-SCAPE clustering to remove highly similar BGCs.
    
    \textbf{MIBiG}---The Minimum Information about a Biosynthetic Gene cluster (MIBiG) database provides manually curated, experimentally validated BGCs with detailed annotations of gene function and final products~\cite{terlouw2023mibig}. We use MIBiG version 3.0, which includes 2,502 entries. This dataset serves as the source for multiple probe tasks in our study. Task-specific subsets are extracted based on criteria such as the presence of halogenase or A-domains.
    
    \textbf{Pfam}---Pfam is a database of protein families, each defined by a conserved domain, a brief functional description, and a unique Pfam ID. The Pfam 33.1 contains around 19,000 domain entries, each associated with a profile HMM used to scan amino acid sequences for domain annotation~\cite{mistry2021pfam}. In this study, Pfam domain-annotated sequences constitute the input to BiGCARP, while the corresponding functional descriptions are used for downstream analyses, including label generation and UMAP-based representation visualization.

\section{Results}
    \subsection{Representation Analysis}

        \subsubsection{UMAP}
         In the original BiGCARP paper \cite{rios2023deep}, the authors present a UMAP projection of ESM-generated embeddings for all Pfam domains in the Pfam database. When colored by Pfam clan membership, the embeddings form coherent clusters, indicating that ESM captures homology-related patterns. This observation motivates the use of ESM-based initialization in BiGCARP, under the assumption that such structure may benefit downstream learning.
         
        However, several limitations should be considered. First, the visualization includes all Pfam domains, whereas only about half are represented in the antiSMASH dataset used to train BiGCARP. When restricting the projection to only those domains, clan-level clustering becomes substantially weaker. Second, clan membership is partially defined by sequence similarity, which ESM embeddings naturally preserve due to their amino acid input. In contrast, BiGCARP is expected to learn higher-level functional patterns beyond sequence similarity. When visualizing the last-layer embeddings of a BiGCARP model trained from random initialization (Fig. \ref{fig:umapbc}), the distribution is largely different, and distinct clusters emerge that align with function labels.
        
        We also compared last-layer embeddings between models initialized randomly and those using ESM-based embeddings. The resulting UMAP projections show similar cluster structures, suggesting that the choice of initialization has limited influence on the final representation space. This explains the comparable downstream performance across initialization schemes, despite differences at the embedding layer.
        
        Due to space constraints, we include one representative visualization (Fig. \ref{fig:umapbc}) in this paper. Additional figures and analyses are available in the accompanying repository.

        \subsection{Functional Themes in UMAP Clusters}

\begin{figure}
            \centering
            \includegraphics[width=.9\linewidth]{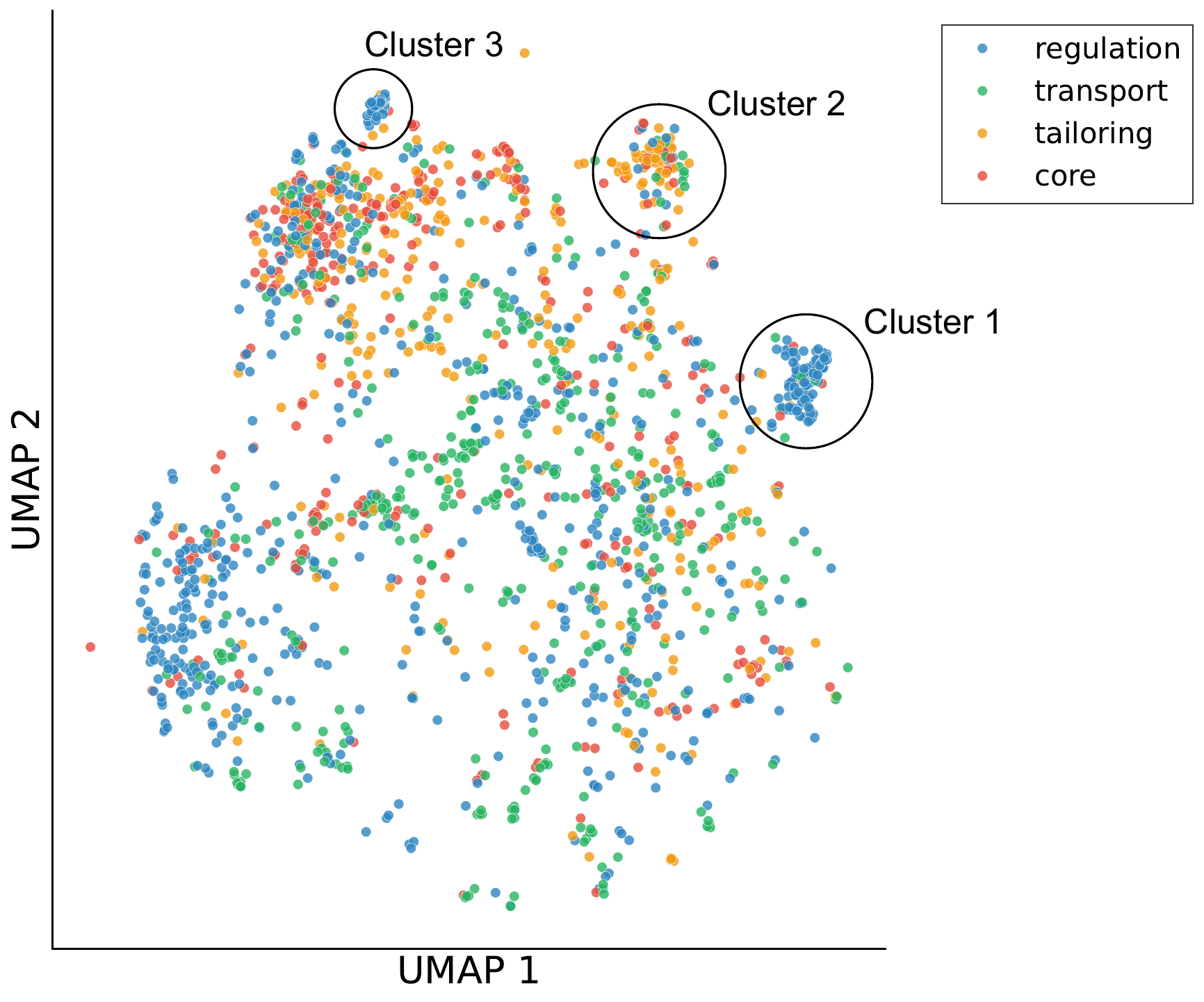}
            \caption{\textbf{Two-dimensional UMAP projection of BiGCARP last-layer embeddings.} All Pfam domains in the training corpus are shown and colored by functional category. Functionally coherent clusters emerge, indicating that BiGCARP captures high-level functional information. Three representative clusters are selected and analyzed in the main text to illustrate their underlying themes.}
            \label{fig:umapbc}
        \end{figure}

        The UMAP visualization revealed several well-defined functional clusters of Pfam domains. Here, we examine three clusters in detail (Fig. \ref{fig:umapbc}).
        
        Cluster 1 consists primarily of regulatory domains involved in the global control of BGC expression. Notable examples include PF00015 (MCPsignal, a methyl-accepting chemotaxis receptor domain) and PF17150 (CHASE-like extracellular sensor domain), both of which detect extracellular or systemic signals and transmit them through signaling cascades to modulate BGC activity. The cluster also contains PF00563 (EAL domain), which degrades cyclic-di-GMP and thereby influences transcription factor binding and pathway activation. Although the underlying mechanisms vary from membrane-bound receptors to second messenger degradation, the unifying theme is global-scale regulatory control of BGCs.
        
        Cluster 2 is enriched in tailoring domains responsible for large-scale modifications of metabolite scaffolds. For example, PF00534 and PF13439 encode glycosyltransferases that append sugar moieties, while PF03062 (a membrane-bound O-acyltransferase) attaches fatty acid chains. These enzymes typically introduce bulky chemical groups, reflecting their role in macro-level structural tailoring.
        
        Cluster 3 contains a mixture of regulatory and tailoring domains operating at a more localized scale. Regulatory elements such as PF00440 and PF08360 (TetR family regulators) often control the expression of nearby genes rather than entire operons. Tailoring domains in this cluster, including GNAT acetyltransferases and PF03492 (SAM-dependent methyltransferases), add small chemical groups like acetyl or methyl moieties, consistent with fine-scale chemical modification.
        
        These clusters demonstrate that the learned embeddings organize domains not only by functional type but also by functional scale: Cluster 1 reflects global regulation, Cluster 2 macro-level tailoring, and Cluster 3 local regulation and fine-scale modification. This separation, despite overlaps in functional labels, suggests that the embeddings capture biologically meaningful contextual information, enabling functional inference for uncharacterized domains in novel BGCs.

        Beyond the main clusters, many domains fall within a diffuse region. We examined several representative cases, such as PF01544 (a general magnesium transporter), and PF14011 (a type VII secretion accessory protein). These examples suggest that the diffuse regions capture broadly reused functions, though they may also reflect limitations of the training data. A systematic analysis will allow drawing firmer conclusions.

        \subsubsection{CKA Analysis of Representation Retention}
        To investigate how pre-trained information propagates through the BiGCARP model, we use CKA to quantify the similarity between internal representations across different layers. Specifically, we aim to assess whether the structural features encoded in the pre-trained ESM embeddings are preserved during training.

\begin{figure}[htbp]
            \centering
            \includegraphics[width=.85\linewidth]{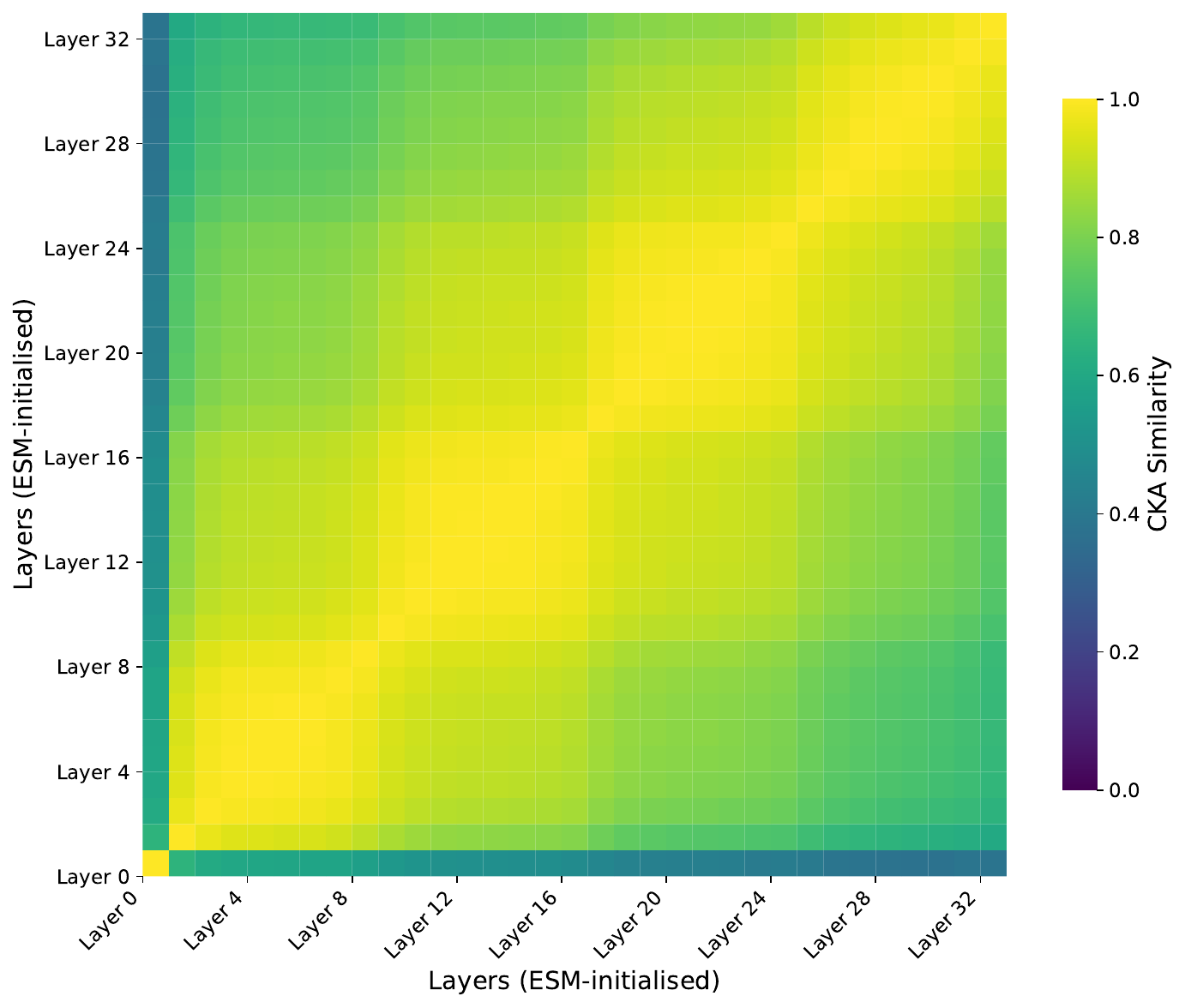}
            \caption{\textbf{CKA self-similarity of the ESM-initialized model across layers.} The four-period pattern reflects the model’s four-block architecture. A sharp change between layers 0 and 1 arises from the dilated convolution layer, while the low similarity between the first and last layers indicates limited retention of initial representations.}
            \label{fig:cka}
        \end{figure}

        Fig. \ref{fig:cka} shows the CKA similarity matrix computed between layers of a BiGCARP model initialized with ESM embeddings. The heatmap reveals four square blocks along the diagonal, reflecting the four-block structure of the CARP architecture. This periodicity serves as a useful consistency check and confirms that layer representations are locally coherent within each block.

        A particularly notable pattern is the sharp drop in similarity immediately after the embedding layer (Layer 0), suggesting that the model rapidly transforms the input embeddings in early stages. This abrupt change is attributable to the first dilated convolution layer, which incorporates local context and substantially modifies the initial representation space.
        
\begin{figure}[htbp]
            \centering
            \includegraphics[width=.9\linewidth]{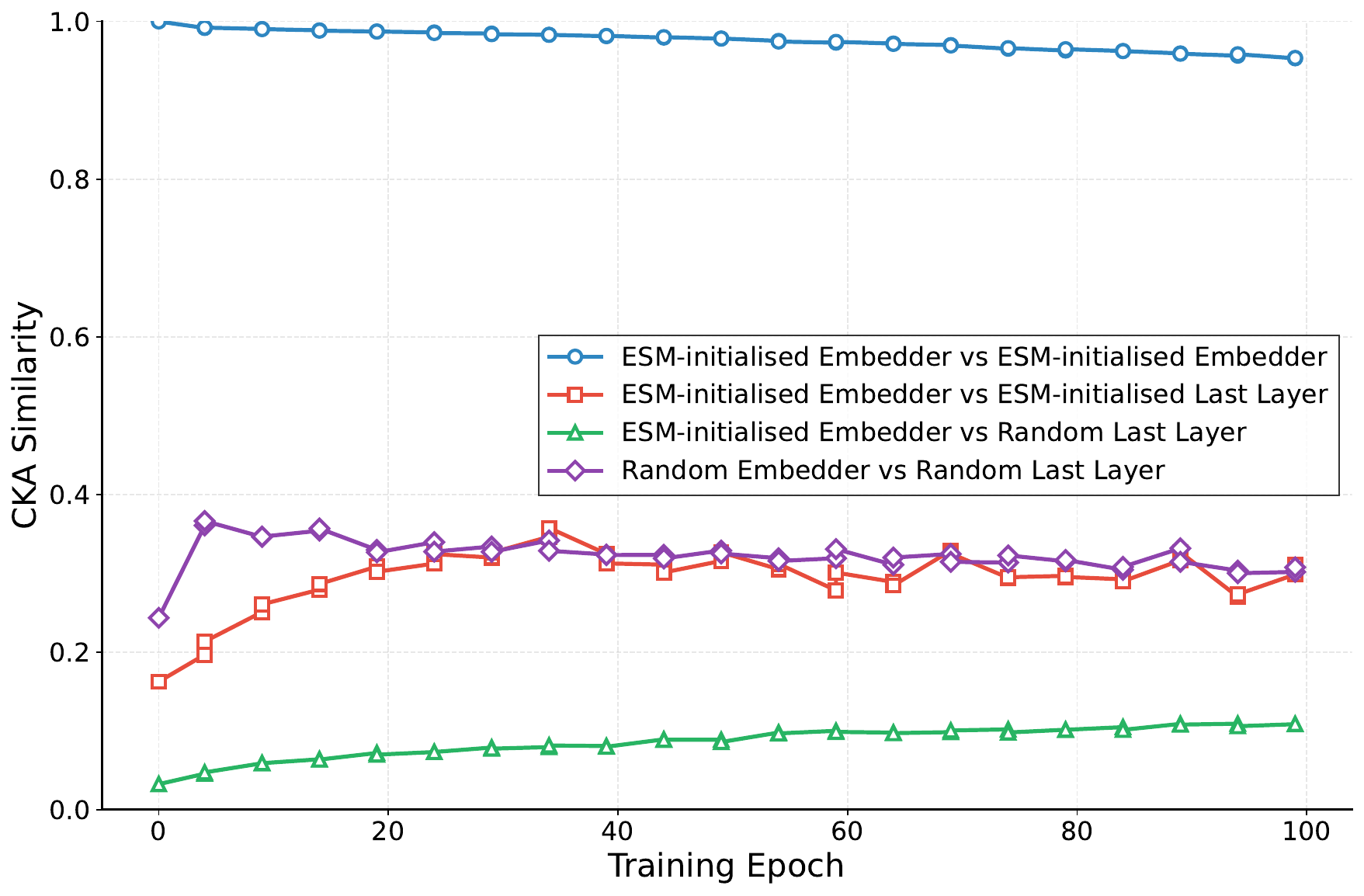}
            \caption{\textbf{Evolution of CKA similarity during training across layers and initialization schemes.} The near-constant similarity of the ESM-initialized embedder to itself (blue) indicates that it is trainable but minimally reshaped during training. The red and purple curves show that, despite different initialization strategies, the similarity between initial and last layers converges to a similar level, suggesting that ESM initialization does not preserve additional information in the final layer. The low similarity of cross-model comparisons (green) provides a consistency check.}
            \label{fig:ckaevo}
        \end{figure}

        To quantify and compare how much of the original embedding structure is retained, we compare the similarity between the embedder and final layers in both ESM-initialized and randomly initialized BiGCARP models throughout the training checkpoints (Fig. \ref{fig:ckaevo}). Interestingly, the CKA curves for both settings follow nearly identical trends, indicating that the information from the ESM initialization is not retained.
        
        This observation suggests that the benefits of pre-trained initialization are largely overwritten under the masked language modeling objective. The final layers in both ESM-initialized and randomly initialized models become highly similar, echoing the findings from the original BiGCARP study that initialization has a limited impact on downstream performance.

    \begin{table}[htbp]
    \caption{BGC Product Class Prediction Performance}
    \begin{center}
    \renewcommand{\arraystretch}{1.2}
    \begin{tabular}{p{6cm}c}
    \toprule
    \textbf{Embedding Strategy} & \textbf{Macro AUROC} \\
    \midrule
    \textbf{ESM + BiGCARP (random-initialized, last-layer)} & $\mathbf{0.917}^{+0.003}_{-0.003}$ \\
    ESM-initialized (last-layer)         & $0.913^{+0.005}_{-0.005}$ \\
    Random-initialized (last-layer)     & $0.912^{+0.005}_{-0.005}$ \\
    Pfam2Vec + Random Forest                    & $0.911^{+0.003}_{-0.003}$ \\
    ESM embeddings                   & $0.908^{+0.003}_{-0.003}$ \\
    Random-initialized (embedder)        & $0.903^{+0.003}_{-0.003}$ \\
    Random baseline                             & $0.889^{+0.005}_{-0.005}$ \\
    ESM-initialized (embedder)            & $0.872^{+0.008}_{-0.008}$ \\
    \bottomrule
    \end{tabular}
    \label{tab:bgcpred}
    \end{center}
    \end{table}

    \subsection{Probe Tasks}
    
    To complement the representation analysis, we use a set of probe tasks to evaluate the quality and characteristics of embeddings produced under different model configurations. These tasks assess the effect of initialization, the impact of embedding extraction methods, and the complementarity between ESM and BiGCARP representations.
    
        \subsubsection{BGC Product Class Prediction}
        The performance of different embeddings is summarized in Table \ref{tab:bgcpred}. Across all configurations, last-layer embeddings from BiGCARP consistently outperform those from the embedder layer and the pre-trained ESM model, irrespective of initialization. In contrast, both ESM and embedder-layer embeddings, which primarily capture local sequence features, perform comparably to the random embedding baseline, indicating their limited utility for this global classification task. Furthermore, the simple concatenation of ESM-based and BiGCARP (random-initialized, last layer) embeddings actually lead to a substantial improvement.
        These results empirically support three main conclusions: (1) BiGCARP’s last-layer embeddings encode meaningful, task-relevant global patterns not present in the initial representations; (2) pre-trained initialization offers little benefit, as training rapidly overrides the initial structure; and (3) combining the cross-granularity embeddings substantially improved the BGC product classification task performance.

        \begin{table}[htbp]
        \caption{A-Domain Substrate Property Prediction Performance (AUROC)}
        \label{tab:adomain}
        \centering
        \renewcommand{\arraystretch}{1.2}
        \begin{tabular}{lccc}
        \toprule
        \textbf{Property} & \textbf{Stachel (ESM)} & \textbf{BiGCARP} & \textbf{BiGCARP + Stachel} \\
        \midrule
        is\_aromatic      & $\mathbf{0.935}^{+0.007}_{-0.009}$ & $0.813^{+0.008}_{-0.010}$ & $0.872^{+0.005}_{-0.015}$ \\
        has\_heterocycle  & $\mathbf{0.955}^{+0.004}_{-0.008}$ & $0.895^{+0.006}_{-0.006}$ & $0.920^{+0.008}_{-0.010}$ \\
        high\_polarity    & $\mathbf{0.913}^{+0.004}_{-0.008}$ & $0.620^{+0.011}_{-0.009}$ & $0.843^{+0.011}_{-0.020}$ \\
        is\_Val           & $\mathbf{0.960}^{+0.003}_{-0.004}$ & $0.909^{+0.004}_{-0.005}$ & $0.937^{+0.008}_{-0.011}$ \\
        is\_Gly           & $\mathbf{0.980}^{+0.002}_{-0.004}$ & $0.923^{+0.004}_{-0.005}$ & $0.958^{+0.008}_{-0.011}$ \\
        is\_canonical\_aa & $\mathbf{0.946}^{+0.005}_{-0.006}$ & $0.885^{+0.005}_{-0.009}$ & $0.927^{+0.007}_{-0.008}$ \\
        \bottomrule
        \end{tabular}
        \end{table}

        \subsubsection{A-Domain Substrate Prediction}
        The A-domain substrate prediction task targets the identification of substrate specificity in adenylation domains, which depends heavily on local sequence features. Therefore, this task serves to evaluate how well the learned embeddings capture locality.

        As shown in Table \ref{tab:adomain}, ESM-based embeddings significantly outperform BiGCARP embeddings, consistent with expectations given ESM’s fine-grained, amino acid-level resolution. While BiGCARP performs worse, its embeddings still achieve well-above-random performance (AUROC = 0.5), indicating that the model retains some useful local context despite being optimized for global representations.
        
        We further evaluated a simple concatenation of ESM and BiGCARP embeddings, but this did not improve performance over ESM alone. This likely reflects the task’s strong reliance on local information, which is already well captured by ESM.
        
        Overall, these results reinforce the complementary nature of the two models: ESM excels at fine-grained, structure-level tasks, while BiGCARP encodes broader, higher-level context more suited to global objectives.

        \subsubsection{Halogen Product Prediction}
        On the halogenation prediction task (Table~\ref{tab:halogen}), concatenating ESM and BiGCARP embeddings achieves the highest AUROC, with a statistically significant improvement over ESM alone ($p = 0.046$, two-sided bootstrap test on LOOCV predictions) and others. This result provides direct evidence that integrating representations across granularities yields measurable benefits for this intermediate-level task.

        The performance pattern aligns with the task’s reliance on finer sequence features: ESM performs strongly due to its residue-level resolution, whereas BiGCARP alone underperforms, indicating that global or local context by themselves are insufficient. The significant gain from simple concatenation suggests that the two models encode complementary information, motivating more expressive fusion strategies and tests on larger annotated datasets.
        
        \begin{table}[htbp]
        \caption{Halogen Prediction Performance (AUROC)}
        \label{tab:halogen}
        \centering
        \renewcommand{\arraystretch}{1.2}
        \begin{tabular}{lc}
        \toprule
        \textbf{Embedding Strategy} & \textbf{AUROC} \\
        \midrule
        \textbf{ESM + BiGCARP (mean-pooled)} & $\mathbf{0.759}^{+0.165}_{-0.209}$ \\
        ESM embeddings                       & $0.659^{+0.216}_{-0.241}$ \\
        BiGCARP (mean-pooled)                & $0.181^{+0.154}_{-0.130}$ \\
        \bottomrule
        \end{tabular}
        \end{table}

\section{Limitations and Future Work}
This proof-of-concept study has several limitations that highlight promising directions for future research. Methodologically, our deliberately simple setup, which involved basic embedding concatenation, static injection, and a simple MLP, could be enhanced with more sophisticated techniques, such as attention-based fusion or joint pre-training, to integrate information more effectively. Furthermore, the study's scope was limited to BGCs, with some tasks constrained by small sample sizes, underscoring the need to scale the approach to larger datasets and extend the cross-granularity paradigm to new biological domains to improve robustness and demonstrate broader applicability. Future work should refine the integration methods and broaden the scope to fully realize the potential of cross-granularity modeling in biological sequence learning.

\section{Conclusion}
The hierarchical, multi-granular structure of biological sequences reflects an intrinsic form of multimodality, offering rich opportunities for cross-granularity representation learning. In this study, we investigated this potential through a case study on two models: ESM at the amino acid level and BiGCARP at the Pfam domain level. Our representational analyses and curated probe tasks revealed that simple embedding transfer across granularities does not retain pre-trained information, while last-layer embeddings capture more meaningful and contextualized features. Moreover, we confirmed that ESM and BiGCARP encode complementary biological information: ESM specializes in fine-grained, local, and structural properties, while BiGCARP captures broader, functional context. Their combination proves effective for intermediate-level tasks, where simple concatenation yields measurable gains. These findings underscore the value of cross-granularity integration as a promising direction for improving the utility and interpretability of biological foundation models.

\bibliographystyle{IEEEtran} 
\bibliography{ref}

\end{document}